# Sugar, Salt & Pepper - Humanoid robotics for autism

Cristina Gena[a], Claudio Mattutino[a], Stefania Brighenti[a], Andrea Meirone[b], Francesco Petriglia[b], Loredana Mazzotta[b], Federica Liscio[b] , Matteo Nazzario[c] , Valeria Ricci[c] , Camilla Quarato[d], Cesare Pecone[d], Giuseppe Piccinni[d]

[a] *Computer Science Dept., University of Turin, corso Svizzera 185, Turin, 10149, Italy*
[b] *Fondazione PAIDEIA Onlus, via Moncalvo 1, Turin, 10131, Italy*
[c] *Intesa Sanpaolo Innovation Center, corso Inghilterra 3, Turin, 10138, Italy*
[d] *Jumple srl, Via Isonzo, 55/2, Casalecchio di Reno, Bologna, 40033, Italy*

**Abstract**
This paper introduces an experimental trial that will take place from February to June 2021, and which will see the use of the Pepper robot in a therapeutic laboratory on autonomies that will promote functional acquisitions in children diagnosed with high functioning autism/Asperger's syndrome.

**Keywords 1**
Assistive robotics, human-robot interaction, autism

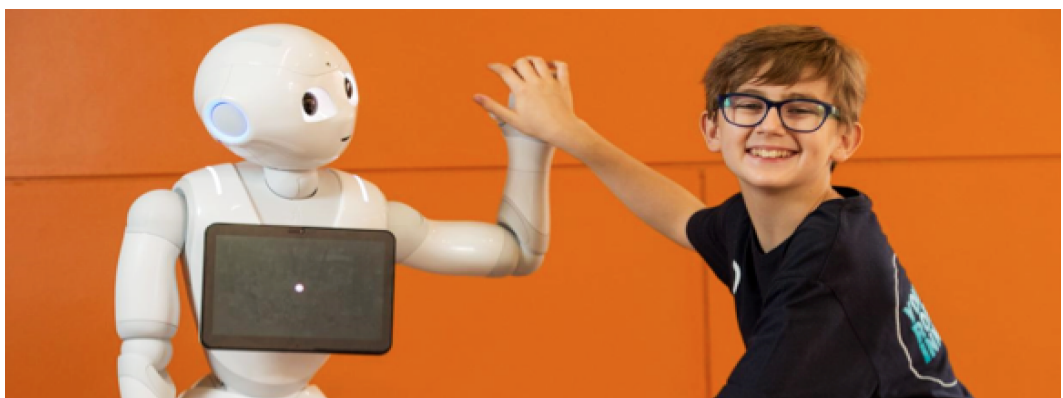

**Figure 1. The Pepper robot.**

## 1. Introduction

Autism spectrum disorders include a group of permanent disabilities that affect the ability to communicate, understand social cues, and recognize and express emotions. Autistic children are generally very attracted to the technology with which they are able to establish social relationships. Following this, since the beginning of the 2000s, research and experiments have been carried out on the use of social robots as therapy tools that have shown how they are particularly engaging and able to arouse new social and communicative behaviors in autistic people, particularly in children and adolescents (Scassellati et al., 2012). Robotic therapy for autism has been explored as one of the first application domains in the field of socially assistive robotics (SAR), which aims to develop robots that help people with special needs through social interaction (Scassellati et al., 2012).

With respect to the specific role of the Pepper robot[2], the reference literature highlights how much children with this diagnosis can be attracted to the robot and implement a greater number of communicative and social initiatives, precisely because the motivation in relating to the robot is higher than others. In the field of Autism Spectrum Disorders, in fact, the interest is directed more towards objects and components of objects and, moreover, there is a particular predisposition and interest in electronic and computer devices. In this sense, Pepper can represent the third party that encourages communicative and social initiative, facilitating the relationship with the other towards which the communicative actions of children must then be gradually directed to. Pepper is placed within what is technically defined *Pairing*, a procedure that can make the other pleasant and motivating for children as it is associated with Pepper (i.e., an adult that initially is considered as a neutral stimulus that becomes a positive stimulus thanks to the contribution of Pepper).

---

[2] https://www.softbankrobotics.com/emea/en/pepper

## 2. Related work

There are a variety of therapeutic approaches that involve the use of robots or other computer technologies in the treatment of autism. (Dautenhahn, 2013). In recent years, several researches have been conducted on the use of social and assistive robots as therapy tools for autism, and these studies have shown that robots are particularly engaging and able to elicit new social and communicative behaviors in people with autism, especially in children and adolescents (Scassellati et al., 2012; Aresti-Bartolome et al., 2014; Pennisi et. al., 2016; Seon-wha, 2013). Social robots are used to develop interventions that help autistic children and to cope with their weaknesses such as recognition and generation of emotions, joint attention and triadic interaction (Chevalier et al., 2016), eye contact and social gaze (Admoni, H ., Scassellati, B, 2017). For example, the minimally expressive humanoid robot KASPAR mediates and encourages interaction between children and co-present adults (Iacono et al., 2011), and Softbank's NAO robot has also been successfully tested in therapy with autistic children with educational games that allow children to work on verbal and non-verbal communication, emotional intelligence, imitation and even basic academic skills (Falconer, 2013).

Furthermore, we mention the importance of educational robotics, a method that allows children to learn thanks to robots. It uses a simple and practical approach to robotics, robot operation, computer programming, visual coding and learning technical subjects such as science and math. Educational robotics is also a method to study STEM subjects (Science, Technology, Engineering and Mathematics) in a practical and fun way, and learn to use logic and solve problems of increasing difficulty that can be formalized (such as a recipe for example), thus forming a computational thinking, which is also one of the objectives of coding. Students with autism prefer this type of problem- solving, where their ability to organize large amounts of data and build reliable structures is useful and produces predictable results (Ribu, 2010). Educational robotics seems particularly interesting for autistic children, who may be focusing on lessons taught in a one-to-one relationship with a robot-teacher and at the same time. Although the topic has been in depth for nearly two decades, very few initiatives have emerged aimed at the use of educational robotics for people with autism (Born, 2011), some of them focus particularly on teaching STEM (Williams, 2018).

## 3. Project main goals

The aim of the research project *Sugar, Salt & Pepper - Humanoid robotics for autism* - envisions the use of the Pepper robot in a therapeutic laboratory on autonomy to promote functional acquisitions in children diagnosed with high functioning Autism Spectrum Disorder / Asperger's syndrome (Level 1 of support according to DSM 5).

We would like also to test the exchanges and interactions of the child with autism in rehabilitation contexts (work on language / communication, on emotions, social skills training, etc.) with the robot to help the operators (psychologists, educators, speech therapists, etc.).

Another objective of the project is to provide young participants with a space within which they can increase their mutual communication and socialization skills (sharing interests, creating friendly bonds, etc.) and strengthening their acquisition of strategies and processes related to daily activities, as snack preparation, kitchen use, school material management, etc., with the aid of the Pepper robot, configured as a highly motivating and scalable tool (also depending on of the user's needs that will be acquired and recorded during the activities and strictly linked to the answers provided by the users and their successful/unsuccessful activities).

Furthermore, we are interested in the collecting and analysing environmental and interpersonal data that can often escape subjective observation but which allow to better outline and deepen the functioning profile of the autistic subject (eye contact, number of communication initiatives, number of requests for help, interpersonal distances, sharing and emotional awareness, etc).

From an computer science point of view, one of the innovative aspects of the research lies in the use of PEPPER together with an external emotion recognition software and a gaze and motion tracker, in order to indirectly analyze the reactions of children to which the robot will be able to adapt [13]. A further innovative aspect concerns the adaptive mechanisms with which PEPPER will be enriched. The robot will show more and more intelligence and reasoning skills, which will allow it to adapt according to the user's needs and customize the interaction to the specific needs of the subject in both adaptive and adaptable form. Thus the robot will become an adaptive robot with respect to the user, able to adapt its behavior based on the characteristics of the subject with whom it is interacting and to customize the interaction in an adaptive form, that is, configurable by the staff expert in therapy.

## 4. Conclusion

In term of expected results we expect the following ones at the end of the experimentation:

- Positive evaluation in terms of perceived experience (user experience and user acceptance) by the participants
- Consolidation or development of effective communication and social strategies
- Consolidation or learning of diversified procedural strategies in a laboratory context
- Transfer of skills learned in life and family contexts
- Collection and analysis of personal and environmental data
- Study and analysis in a real and longitudinal context (in the wild evaluation) of a child-robot application, in the research field of social and assistive robotics
- Study and analysis and refinement of adaptive strategies based on affective computing and biometrics (emotions, gaze, gestures, etc.)
- Study and analysis of end-user development interfaces for robot programming and behavior

## 5. Acknowledgements

This project has been funded by Intesa Sanpaolo Bank - Bank of the Territories Division.